\documentclass{article}
\usepackage{spconf,amsmath,epsfig}

\let\OLDthebibliography\thebibliography
\renewcommand\thebibliography[1]{
  \OLDthebibliography{#1}
  \setlength{\parskip}{0pt}
  \setlength{\itemsep}{0pt plus 0.3ex}
}

\usepackage{amsmath}
\usepackage{amssymb}

\usepackage{times}
\usepackage{epsfig}
\usepackage{graphicx}

\usepackage{multicol}

\usepackage{caption,subcaption}

\usepackage{gensymb}
\usepackage{float}
\usepackage{subcaption}
\usepackage{multirow}

\usepackage[utf8]{inputenc}
\usepackage[english]{babel}
\usepackage[normalem]{ulem}

\usepackage{soul}
\usepackage[norule,symbol,perpage]{footmisc}

\usepackage[table,xcdraw]{xcolor}

\usepackage{booktabs}
\usepackage{booktabs}
\usepackage{graphicx}

\newcommand{\etal}{\textit{et al}. } 

\usepackage{soul}

\begin{document}\sloppy
\topmargin=0mm
\def\x{{\mathbf x}}
\def\L{{\cal L}}

\title{Identifying the Origin of Finger Vein Samples \\Using Texture Descriptors}

\twoauthors{Babak Maser$^{\dagger}$\thanks{$^{\dagger}$ORCID iD: 0000-0002-1662-8324}}
{Multimedia Signal Processing \\ \& Security Lab$^{\ddagger}$\thanks{$^{\ddagger}$http://www.wavelab.at/index.shtml} \\
University of Salzburg, AUSTRIA\\
babak.maser@stud.sbg.ac.at}
{Andreas Uhl}
{Multimedia Signal Processing \\ \& Security Lab$^{\ddagger}$ \\
University of Salzburg, AUSTRIA\\
Uhl@cosy.sbg.ac.at}


\maketitle


%
\begin{abstract}

\textit{Identifying the origin of a sample image in biometric systems can be beneficial for data authentication in case of attacks against the system and for initiating sensor-specific processing pipelines in sensor-heterogeneous environments. Motivated by shortcomings of the photo response non-uniformity (PRNU) based method in the biometric context, we use a texture classification approach to detect the origin of finger vein sample images. Based on eight publicly available finger vein datasets and applying eight classical simple texture descriptors and SVM classification, we demonstrate excellent sensor model identification results for raw finger vein samples as well as for the more challenging region of interest data. The observed results establish texture descriptors as effective competitors to PRNU in finger vein sensor model identification.}
\end{abstract}
\section{Introduction}
\label{sec:intro}

Nowadays we encounter a significant surge in the use of unattended applications of biometric systems. In many biometric modalities, a digital biometric image sensor is the core component for data acquisition, operating in the near-infrared (NIR) domain in the case of finger vein recognition.

Deducing sensor information from the images serves as a basis for different forensic and non-forensic tasks. One of the major tasks in digital image forensics is establishing an image’s origin with the help of the deduced sensor information. This can be performed at different levels: Sensor-technology, brand, model, unit. In the context of biometric systems the extracted sensor information can be used for various applications. In this work we focus on two specific ones: Securing a finger vein recognition system against insertion attacks and enabling device selective processing of the image data.

The authenticity and integrity of the acquired biometric sample data plays an important role for the overall security of the biometric system,
in particular in the case of unattended operation. Among other attacks (e.g. presentation attacks which present spoofing artefacts to the biometric sensor), an {\it insertion attack} bypasses the biometric sensor by inserting data (i.e. biometric samples) into the transmission from the sensor to the feature extractor / comparison module. The finger vein image inserted during the attack could have been  acquired with another sensor off-site, even without the knowledge of a genuine user, or could be a manipulated image to spoof the biometric recognition system. 

In large-scale biometric system various sensors from different manufacturers and models are deployed and the interoperability is often affected by specifics of each sensor, such as the acquisition technique or in-sensor image processing. Selective processing of the images helps to improve the interoperability by applying a sensor tailored biometric toolchain. Therefore information about the sensor model is required, which can be deduced from the iris images directly utilising image forensic methods.

This work evaluates the feasibility of deducing sensor information at model level, i.e. the biometric sensor a finger vein sample image is captured with, from the finger vein image using image texture based methods. 
To learn an image's origin, various methods have been proposed, the approach exploiting the photo response non-uniformity (PRNU) being the most prominent one \cite{chen2008determining}, as it also allows a sensor identification at unit level.

However, the application of PRNU-based techniques in biometric sample data  authentication has exhibited some difficulties: First, the PRNU fingerprint can be extracted from images of a biometric sensor and injected into forged sample images \cite{goljan2010sensor,Uhl12d}. Only under certain restrictive conditions such attacks can be detected or avoided. Second, authentication results with respect to biometric sensors have been reported to be widely varying at best (see e.g.
\cite{kalka2015preliminary,Uhl12d} for iris sensor identification) and have been
shown to be dependent and influenced by sensor components depicted in the 
images (see e.g. \cite{8987237} for finger vein sensor identification). One reason for the difficulties is the requirement to compute the PRNU fingerprint from uncorrelated data - which is of course hard to satisfy given the high similarities among sample images present in biometric datasets \cite{Debiasi14a}. Attempts to clarify this issue have not been convincing so far \cite{Debiasi15a}.
As a consequence, texture classification techniques have been proposed to identify iris sensors at model level earlier \cite{debiasi2017identifying,el2015dataset}. However, contrasting to this work, the authors of the former work propose rather costly techniques, i.e. improved Fisher vector encoding of denseSIFT and dense Micro-block difference features.




        

	\begin{figure*}
	    \centering
	    \fbox{
		    \begin{minipage}{.249\textwidth}
	        \vspace*{3mm}
	        \centering
	        \includegraphics[width=1.0\textwidth]{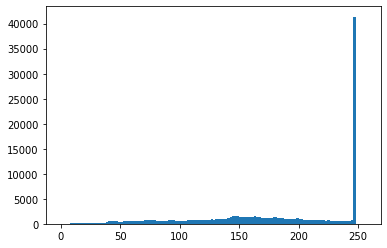}
	        \vspace*{5mm}
	        \includegraphics[width=4cm, height=1.8cm]{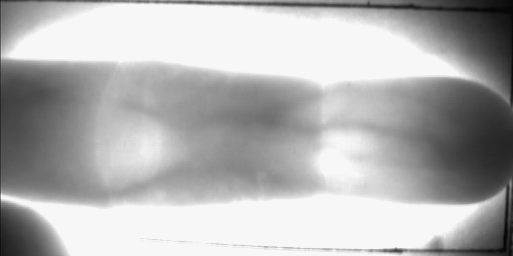}
	        \subcaption{}
	    \end{minipage}%
	    \begin{minipage}{.249\textwidth}
	        \vspace*{5mm}
	        \centering
	        \includegraphics[width=1.0\textwidth]{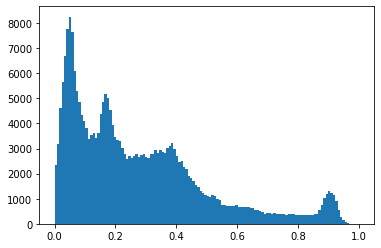}
	        \vspace*{5mm}
	        \includegraphics[width=4.0cm, height=1.8cm]{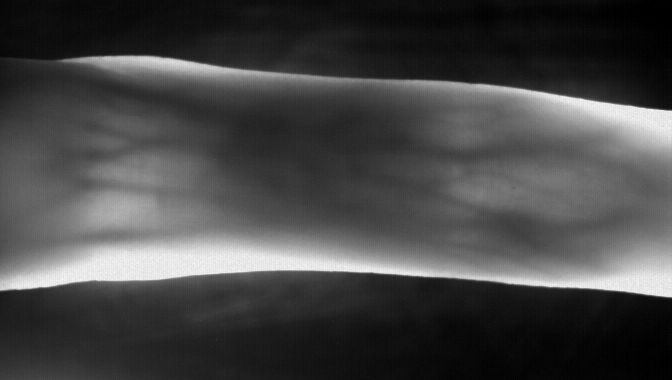}
	        \subcaption{}
	    \end{minipage}%
	    \begin{minipage}{.249\textwidth}
	        \vspace*{5mm}
	        \centering
	        \includegraphics[width=1.0\textwidth]{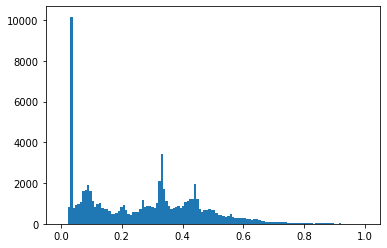}
	        \vspace{5mm}
	        \includegraphics[width=4.0cm, height=1.8cm]{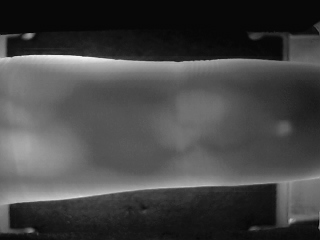}
	        \subcaption{}
	    \end{minipage}%
	    \begin{minipage}{.249\textwidth}
	        \vspace*{-5mm}
	        \centering
	        \includegraphics[width=1.0\textwidth]{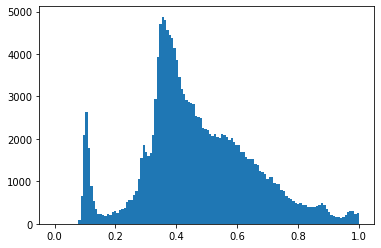}
	        \vspace*{1mm}
	        \includegraphics[width=4.0cm, height=1.8cm]{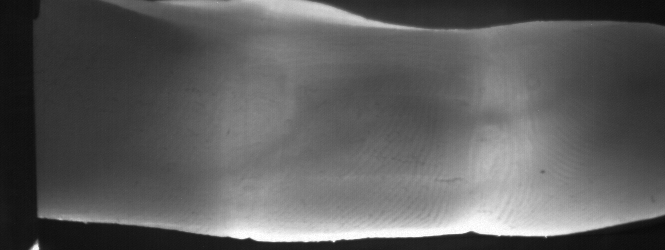}
	        \subcaption{}
	        \vspace*{-7mm}
	    \end{minipage}
			}
	    \caption{ Image and corresponding histogram  samples of original sample images of (a)\hspace{1pt} HKPU\_FV dataset, (b)\hspace{1pt} UTFVP dataset, (c)\hspace{1pt} SDUMLA dataset, and, (d)\hspace{1pt} IDIAP dataset.}
	    \label{fig:histograms}
		
	\end{figure*}


This work is structured as follows: In section \ref{SubSec:datasets} we discuss
the properties of the finger vein sample datasets as considered.
Section \ref{sec:experimental_design} explains the conducted experiments in depth, where subsection \ref{Sec:methodology} describes the used texture description methodology. Next, We discuss and analyze the experimental results in section \ref{sec:result}, and finally, we end this manuscript with a conclusion in section \ref{sec:conclution}.

\section{Fingervein Sample Data} 
\label{SubSec:datasets}
 We considered eight different public finger vein datasets (acquired with distinct prototype near infrared sensing devices), taking 120 samples from each dataset. As in finger vein recognition features are typically not extracted from a raw sample but from a region-of-interest (ROI) image containing only finger vein texture, 
 an insertion attack can also be mounted using such ROI data (in case the sensor does not deliver a raw sample to the recognition module but ROI data instead). Thus, we produced cropped ROI datasets out of the original ones (description of methodology is given afterwards) to be able to test these data for their distinctiveness as well. Subsequently, we briefly detail the specifications of each dataset: 

\begin{itemize}
    \item \textbf{SDUMLA-HMT \cite{BYin11a}:} Original resolution is 240$\times$320, ROI data is 85$\times$320 pixel. 120 images are selected from the first 30 subjects. 
    
    \item \textbf{HKPU-FV \cite{BKumar12a}:} Original resolution is   256$\times$513,  ROI data is 60$\times$390 pixel. 120 images are selected from the first 60 subjects. 
    
    \item \textbf{IDIAP \cite{Tome_IEEEBIOSIG2014}:} Original resolution is 250$\times$665,  ROI data is  125$\times$610 pixel.  120 images are selected   from the first 60 subjects.

    \item \textbf{MMCBNU\_6000 (MMCBNU) \cite{BLu13a} :} Original resolution is 640$\times$480, ROI data is 155$\times$620 pixel. 120 images are selected  from the first 20 subjects. 
    
    \item \textbf{PLUS-FV3-Laser-Palmar (Palmar) \cite{Kauba18c}:} Original resolution is 600$\times$1024, ROI data is 110$\times$500 pixel. 120 images are selected from the first 20 subjects. 
    
    \item \textbf{FV-USM \cite{BAsaari14a}:} Original resolution is 480$\times$640 pixels, ROI data is 110$\times$280 pixel. 120 images are selected from the first 30 subjects.  
     
    \item \textbf{THU-FVFDT  \cite{Kauba18c}:} Original resolution is 600$\times$1024, ROI data is 120$\times$390 pixel.  120 images are selected  from the first 120 subjects.
    
    \item \textbf{UTFVP \cite{BTon13a}:} Original resolution is 380$\times$672, ROI data is 140$\times$490 pixel. 120 images are selected from the first 60 subjects.
\end{itemize}
     
Original sample images, as shown in Fig. \ref{fig:histograms}, can be discriminated easily: Besides the differences in size (which can be adjusted by an attacker of course), the sample images can be probably distinguished by the extent and luminance of background. To illustrate this, we display the images' histograms above each example in Fig. \ref{fig:histograms}, and those histograms
clearly exhibit a very different structure. Thus, we expect texture descriptors
to have an easy job to identify the origin of the respective original sample images.
   
\subsection{Generating Region of Interest Datasets}
\label{subsec:cropped_datasets}

In finger vein recognition, contrasting to e.g. fingerprint recognition,
feature extraction is not applied to the entire raw sample data but instead
to a ROI only \cite{Kauba18c,BKumar12a}. In this ROI, only actual finger texture is contained. Depending on the setup of the system, the sensor might already extract the ROI from the raw sample. In this setting, identification of the 
finger vein data's origin has to be based on the ROI, thus, it will be required under these circumstances that only finger vein texture (the ROI) is used to discriminate sensors. This is not unrealistic, as in the iris recognition case, normalised iris texture has been considered by analogy to be used for sensor identification \cite{debiasi2017identifying,el2015dataset} instead of raw iris sample data.

\begin{figure}
    \fbox{
    \centering
    \begin{subfigure}[t]{.45\textwidth}
    \vspace{2pt}
    \centering
        \raisebox{-\height}{\includegraphics[width=0.30\textwidth]{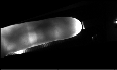}}%
        \hspace{1pt}
        \raisebox{-\height}{\includegraphics[width=0.30\textwidth]{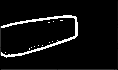}}%
        \hspace{1pt}
        \raisebox{-\height}{\includegraphics[width=0.30\textwidth]{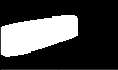}}%
        \vspace{.2ex}
        \raisebox{-\height}{\includegraphics[width=0.30\textwidth]{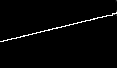}}%
        \hspace{1pt}
        \raisebox{-\height}{\includegraphics[width=0.30\textwidth]{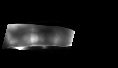}}%
        \hspace{1pt}
        \raisebox{-\height}{\includegraphics[width=0.30\textwidth]{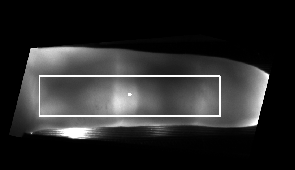}}
    \end{subfigure}
    }
     \caption{ROI generation for datasets with finger position variability.}
     \label{fig:StepsToCropSampleImage}
\end{figure}

To detect and segregate the finger vein region and extract a patch consisting of biometric data only, different techniques have been applied depending on the properties of each dataset. For the datasets exhibiting a higher intra-variance of finger positions, we applied the following algorithm based on
morphological snakes (morphological active contour without edges \cite{902291})  to extract the ROI (FV\_USM, THU\_FVDT, UTFVP, and MMCBNC datasets), also illustrated in Fig. \ref{fig:StepsToCropSampleImage}:

 \begin{enumerate}
     \item Apply morphological snakes to the finger vein image to produce a segmented image.
     \item Apply Canny edge detection and contour closing to detect the finger vein region.
     \item Fill the contour.
     \item Find the mass center of the filled contour and fit a line to the contour; estimate the angle ($\theta$) of the line to the x-axis.
     \item Rotate the texture area by $\theta$ degree.
     \item Find the new mass center and crop the aligned original sample image. 
 \end{enumerate}

For the Palmar dataset we also used this method but replaced the morphological snakes technique by the Chan-Vese segmentation algorithm \cite{getreuer2012chan}.

To create the ROI for the (easier) datasets HK\_FV, IDIAP and SDUMLA we applied the following steps:

 \begin{enumerate}
     \item Apply Canny edge detection.
     \item Apply a dilation operator on the detected edges. 
     \item Stack all images of a dataset on top of each other.
     \item Extract the common patch patch of finger vein texture.
 \end{enumerate}

\begin{figure}[h!]
    \centering
 
    \includegraphics[width=0.49\linewidth]{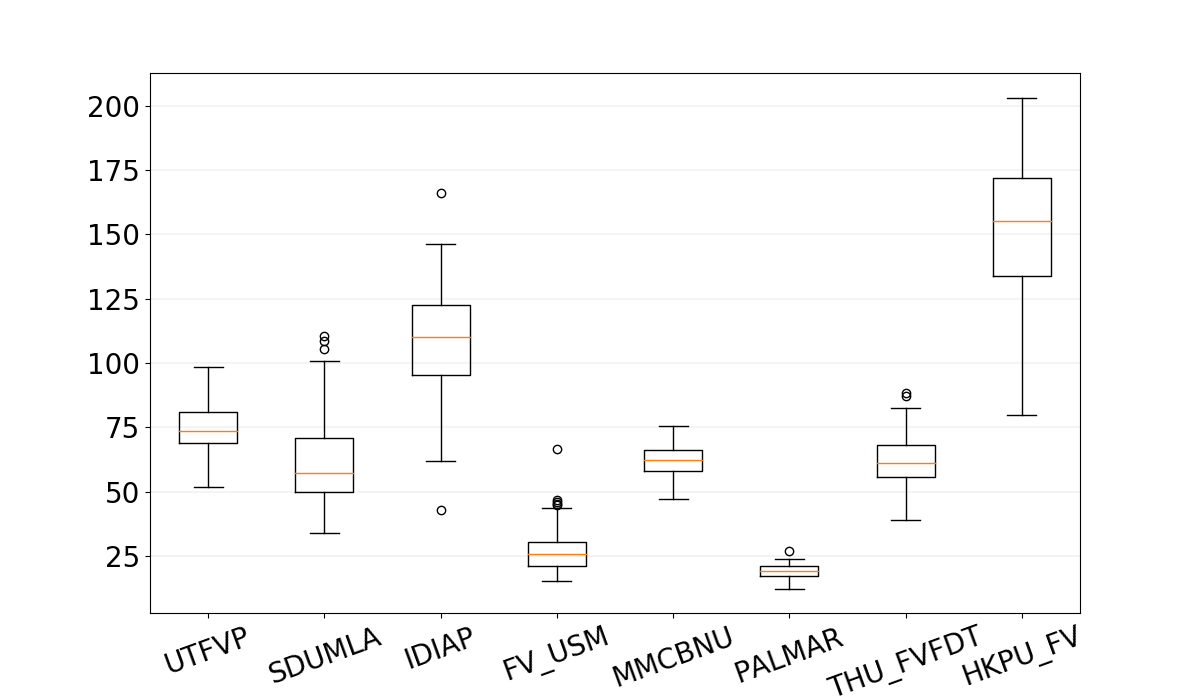}
    \includegraphics[width=0.49\linewidth]{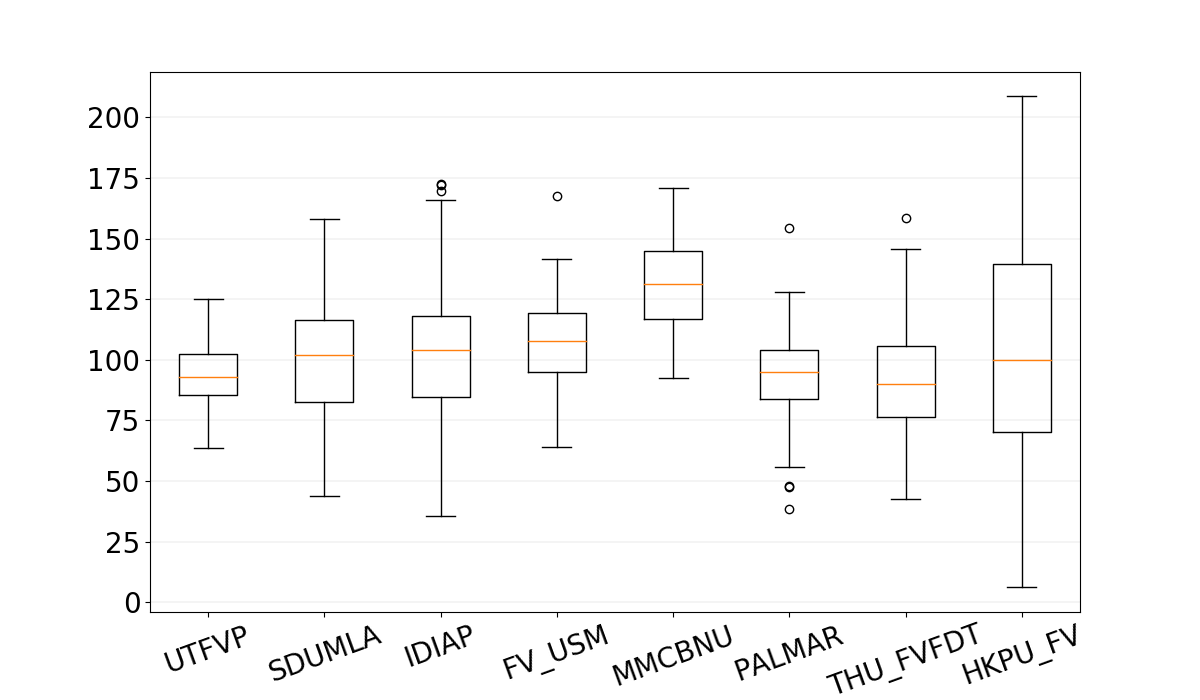}
    \caption{Luminance distribution of original and ROI images across all datasets, respectively.}
   
    
    \label{fig: luminance_uncropped}
\end{figure}

\begin{figure}[h!]
    \centering
    \includegraphics[width=0.49\linewidth]{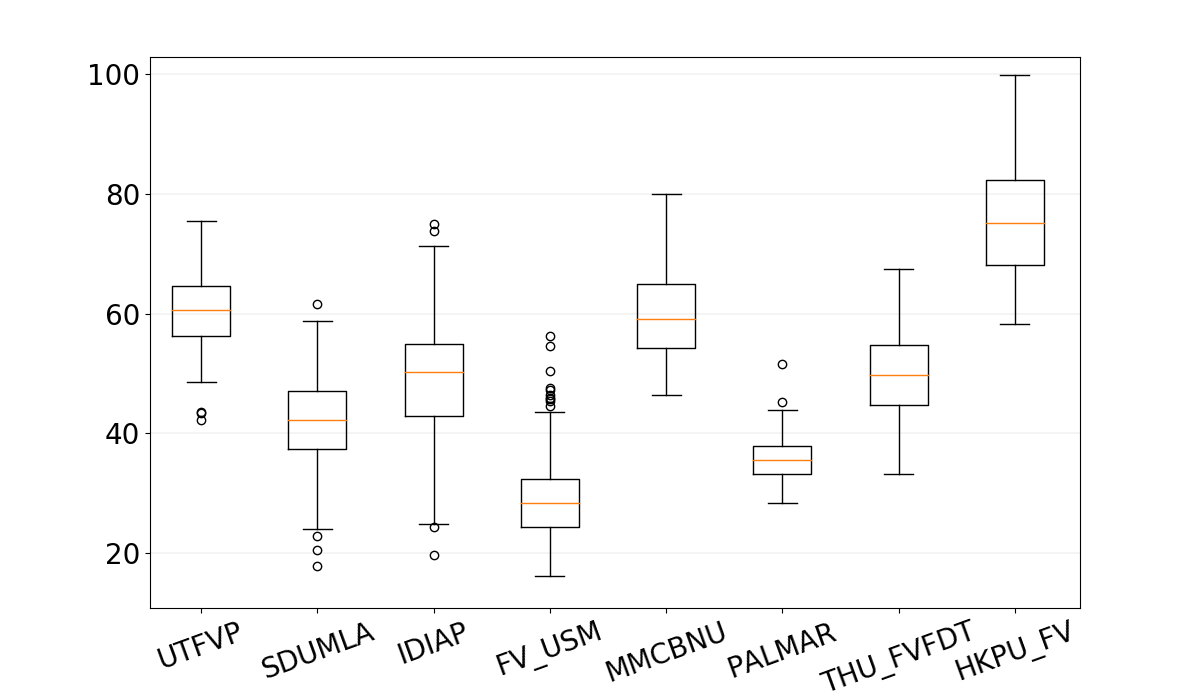}
    \includegraphics[width=0.49\linewidth]{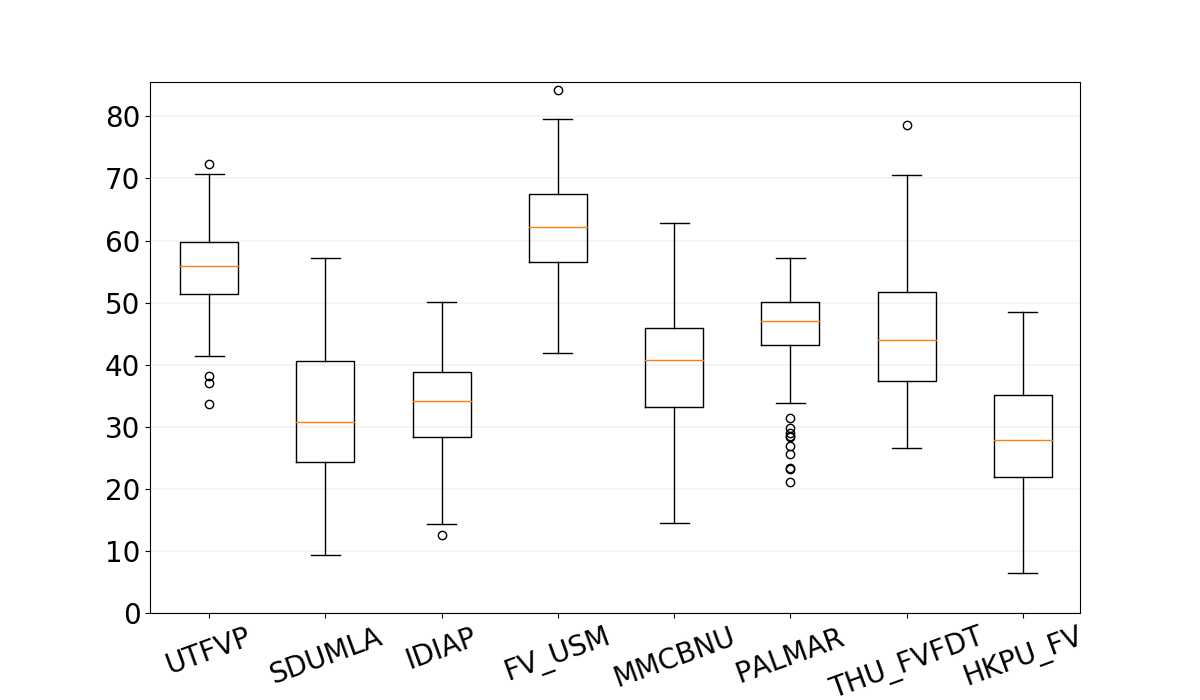}
    \caption{Variance distribution of original and ROI images across all datasets, respectively.}
    
    \label{fig:variance_uncropped}
\end{figure}

\begin{figure*}
    \centering
    \fbox{%
    \begin{minipage}{.3\textwidth}
        \vspace{5mm}
        \centering
        \includegraphics[width=0.7\textwidth]{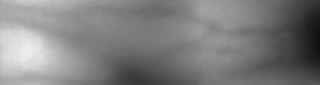}
        \subcaption{SDUMLA ROI}
        \includegraphics[width=0.7\textwidth]{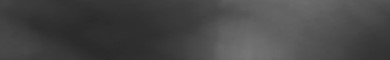}
        \subcaption{HKPU\_FV ROI}
        \includegraphics[width=0.7\textwidth]{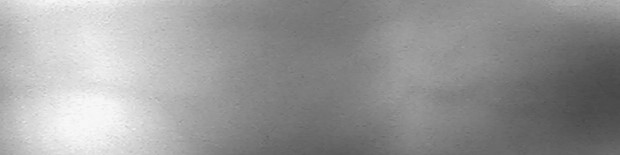}
        \subcaption{MMCBNU ROI}
        \vspace{2mm}
    \end{minipage}%
    \begin{minipage}{.3\textwidth}
        \vspace{5mm}
        \centering
        \includegraphics[width=0.7\textwidth]{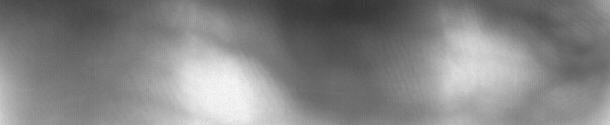}
        \subcaption{IDIAP ROI}
        \includegraphics[width=0.7\textwidth]{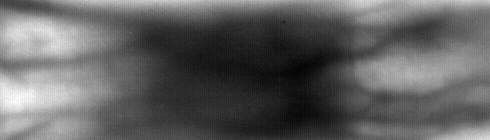}
        \subcaption{UTFVP ROI}
        \includegraphics[width=0.7\textwidth]{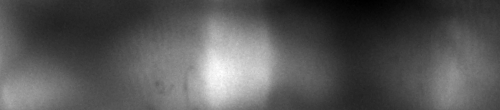}
        \subcaption{PALMAR ROI}
        \vspace{2mm}
    \end{minipage}%
    \begin{minipage}{.3\textwidth}
        \centering
        \includegraphics[width=0.7\textwidth]{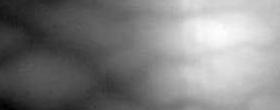}
        \subcaption{FV\_USM ROI}
        \includegraphics[width=0.7\textwidth]{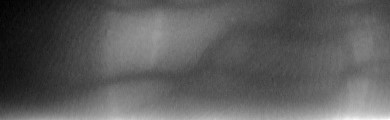}
        \subcaption{THU\_FVFDT ROI}
    \end{minipage}
    }
    \caption{ ROI Samples from different datasets}
  \label{ROIs}  
\end{figure*}

Fig. \ref{ROIs} illustrates the results of ROI creation for a sample of each dataset (sample width has been normalised for better clarity). It gets immediately clear that discrimination is obviously more difficult based on the ROI data only. To investigate the differences between raw sample data and ROI data in more detail, we have investigated the range of luminance values and their variance across all datasets. Figs. \ref{fig: luminance_uncropped} and
\ref{fig:variance_uncropped} display the results in the form of box-plots, where
the left box-plot corresponds to the original raw sample data, and the right one to the ROI data, respectively.
We can clearly see that the luminance distribution properties have been changed dramatically once we change our focus from original datasets to ROI datasets. For example, original HKPU\_FV samples can be discriminated from FV\_USM, MMCBNUm, PALMAR, UTFVP, and THU\_FVFDT ones by just considering luminance value distribution. For the ROI data, the differences are not very pronounced any more. When looking at the variance value distributions, we observe no such strong discrepancy between original sample and ROI data, still for some datasets variance can be used as discrimination criterion (e.g. Palmar vs. HKPU\_FV in original data,
FV\_USM vs. HKPU\_FV in ROI data). Consequently, we expect the discrimination of the considered datasets based on texture descriptors to be much more challenging when focusing on the ROI data only.

The steps to produce the cropping images is shown in Fig-\ref{fig:StepsToCropSampleImage}. The size of cropped images for each dataset has been given in the subsection \ref{SubSec:datasets}.



\section{Experimental Design} 
\label{sec:experimental_design}

\subsection{\textbf{Texture Description Methodology}}
\label{Sec:methodology}

To discriminate sensors we applied a number of classical yet simple approaches to produce a texture descriptor of a finger vein image. In the following subsections, we briefly describe the chosen techniques and explain how to cope with differently sized images.

\subsubsection{\textbf{Fourier Ring Filter (FRF)}}
\label{SubSec:fourier_ring_filter}

We generate features in the frequency domain using 2-D FFT. Independent of image size, fifteen band pass filters split the frequency domain into equally sized bands which are used to compute mean and standard deviation of each ring \cite{Vecsei09a} (which are used as statistical texture descriptors).  

\subsubsection{\textbf{Local Binary Patterns (LBP)}}
\label{SubSec:histogram_lbp}

We use a variant of the original \textbf{Local Binary Pattern (LBP)} introduced by Ojala \etal \cite{ojala1996comparative}. This approach is called Histogram-LBP (HLBP \cite{boulogne2014scikit}), we set the radius to 3 and the number of curricular neighborhood pixels is set to 15.
The HLBP is invarant to image size if the output of the histogram for each image is normalized. Further, the number of histogram bins is fixed.

Additionally, we apply uniform LBP (ULBP) -  a LBP is called uniform if the binary pattern contains at most two 0-1 or 1-0 transitions, and it has been shown that these pattern occur more frequently in natural texture (and significantly reduces feature length vectors as the LBP histogram bins are reduced).

\subsubsection{\textbf{Image Histogram (IMHIST) }}
\label{SubSec:histogram}
We simply compute the image histogram and take the output as feature vector. The IMHIST is invariant to image size by bin entry normalisation and fixing the number of histogram bins. 

\subsubsection{\textbf{Wavelet-based Features}}
\label{SubSec:wavelet_mean_variance}

We apply 2-D wavelet decomposition using Daubechies 8-tap orthogonal filters to generate the coefficients in horizontal $h$, vertical  $v$, and diagonal $d$ directions. On every decomposition level we compute mean ($\mu$) and standard deviation (std) for each of the sub-bands $v$, $h$, and $d$ and concatinate those to get the mean and variance feature (WMV).
 We achieved invariance to image size by fixing the number of wavelet decomposition levels to 3.

Similar to WMV, we define wavelet variance (WV) by computing the variance per subband, and wavelet entropy (WE) by computing entropy per subband, respectively.

\subsubsection{\textbf{Local Entropy (LE)}}
\label{SubSec:wavelet_local_entropy}

We slice a given image into 16 blocks (tiles) and compute the entropy from each tile. By taking a histogram of all produced entropies a feature vector is generated. LE also is invariant to image size by fixing the number of image blocks (tiles) and bin number of the histogram.

\subsection{\textbf{SVM Finger Vein Texture Classification (FVTC)}}
\label{subsec:classifier}

 The SVM classifier is trained by feeding $67\%$ of all images then the remaining images $33\%$ are used for the testing purpose. Images in all finger vein datasets are randomly shuffled beforehand to avoid subject-related bias. To optimize the SVM classifier and to obtain the most promising hyperparameters such as $C$, $\gamma$, $kernel$ and $degree$, we employed a Grid Search technique in combination with 4-fold cross-validation \cite{kohavi1995study}. Also, we set the decision function to $"one\; vs\; rest"$ strategy.      

\subsection{\textbf{Evaluation metrics}}
\label{subsec:evaluation_metrics}
We use classical measures to rate our sensor identification task, which is basically a multi-class classification problem.
The multi-class problem is an extension of binary classification. We use two approaches for evaluation: First, receiver operating characteristic (ROC) which relates the false positive rate to the false negative rate, and second, the relation of precision and recall. For both relations, the Area Under The Curve (AUC) can be computed as a single measure. 

Once we have multi-class problem, the challenging point is how to get an overall score. Often, one simply takes the average of the AUC ROC metrics. For illustration, to calculate Recall for three-class problems, we sum up three Recalls and divide them by number of contributing classes, that is classical average. In contrast, another approach is based on summing up individual terms during the computation. The former is called macro-average and latter is called micro-average approach, respectively \cite{ferri2003volume,van2013macro}.

In this paper we use the micro-average approach to estimate the average metrics of our multi-class problem. We use the micro-average AUC ROC (mA AUC ROC) and micro-average AUC Precision-Recall(mA AUC Pr-Re) as our perfomance metrics. In the multi-class setting, we need to estimate the aforementioned metrics by computing e.g. micro-average FPR (False Positive Rate), micro-average Recall, and micro-average Precision as follow:
   
\begin{equation}
\centering
    \begin{split}
        mA\;Pr & = \frac{\sum_c TP_c}{\sum_c TP_c + \sum_c FP_c} 
    \end{split}
\end{equation}%
\begin{equation}
\centering
    \begin{split}
        mA\;Re & = \frac{\sum_c TP_c}{\sum_c TP_c + \sum_c FN_c} 
    \end{split}
\end{equation}%
\begin{equation}
\centering
    \begin{split}
        mA\;FPR & = \frac{\sum_c FP_c}{\sum_c FP_c + \sum_c TN_c} 
    \end{split}
\end{equation}

\vspace*{5mm}

Where c is the class label, TP is True Positive, FN is False Negative, TN is True Negative, and FP is False Positive.\\
In this work, the number of images in the contributed datasets is balanced. Therefore, the value of macro-average and micro-average are very close and sometimes even identical.

\subsection{\textbf{Enhancement techniques}}
\label{subsec:enhancement_techniques}

To enhance the sample images, we applied the following methods.
\begin{enumerate}
    \item \textbf{Wiener Filter and CLAHE (Enh.):} To enhance the quality of the images and remove undesired noise-related artifacts, we apply a Wiener Filter \cite{benesty2005study} and also to improve the contrast of images, we use CLAHE (Contrast Limited Adaptive Histogram Equalization). Applying these two filters is done sequentially on all images of the mentioned datasets.
    
    \item \textbf{No Enhancement (NoEnh.):} Sample images are used as present in the datasets or as obtained after ROI computation.
    
\end{enumerate}

\section{\textbf{Results}}
\label{sec:result}

Table \ref{tbl:scores_of_texture_descriptor_methods} displays the experimental results. As expected, sensor identification is easily achieved based on original samples.  Image enhancement improves results (mostly slightly)
in many cases, we get values $>$ 0.99 in terms of AUC ROC and AUC Pr-Re
for ULBP and WMV for both enhancement settings, which is a perfect result. 

\begin{table}[h!]
\renewcommand\arraystretch{1.5}
    \begin{center}
    \resizebox{0.50\textwidth}{!}{%
    \begin{tabular}{cccccc}
   
    \multicolumn{1}{l}{\textbf{}}   &   
    \multicolumn{1}{l}{\textbf{}}   & \multicolumn{2}{c}{\textbf{Original Sample}} &    \multicolumn{2}{c}{\textbf{ROI}} \\
    \cline{3-6}  
    \multicolumn{1}{c}{\textbf{}} &
      \multicolumn{1}{l}{\textbf{Descriptor}} &
      \multicolumn{1}{c}{\textbf{No.Enh}} &
      \multicolumn{1}{c}{\textbf{Enh.}} &
      \multicolumn{1}{c}{\textbf{No.Enh}} &
      \multicolumn{1}{c}{\textbf{Enh.}} \\ \hline  \hline 
    
    \multicolumn{1}{c}{\textit{\textbf{mA AUC ROC}}}   &  FRF   & 0.997   & 0.999   & 0.989   & 0.986           \\ 
    \multicolumn{1}{c}{\textit{\textbf{mA AUC Pr-Re}}} &     & 0.987   & 0.999   & 0.952    & 0.932          \\ \hline
    
    \multicolumn{1}{c}{\textit{\textbf{mA AUC ROC}}}   &  HLBP   & 0.992    & 0.994   & 0.941   & 0.955           \\ 
    \multicolumn{1}{c}{\textit{\textbf{mA AUC Pr-Re}}} &      &0.960	&0.968	&0.762	&0.783         \\ \hline 
    
    \multicolumn{1}{c}{\textit{\textbf{mA AUC ROC}}}   & ULBP   &0.995	&0.999	&0.931	&0.932          \\ 
    \multicolumn{1}{c}{\textit{\textbf{mA AUC Pr-Re}}} &   & 0.994	& 0.997	& 0.763	& 0.728    \\ \hline 
    
    \multicolumn{1}{c}{\textit{\textbf{mA AUC ROC}}}   & LE & 0.919  & 0.952 & 0.834  &  0.858   \\ 
     \multicolumn{1}{c}{\textit{\textbf{mA AUC Pr-Re}}} &   & 0.685  & 0.772 &	0.538  & 0.650	 \\ \hline

    \multicolumn{1}{c}{\textit{\textbf{mA AUC ROC}}} & ImHist    & 0.989 &	0.961 &	0.906 &	0.966    \\ 
     \multicolumn{1}{c}{\textit{\textbf{mA AUC Pr-Re}}} &   & 0.883 &	0.789 &	0.481 &	0.851       \\ \hline
    
    \multicolumn{1}{c}{\textit{\textbf{mA AUC ROC}}}   &  WV & 0.998	& 0.998	& 0.984	 & 0.983        \\
    \multicolumn{1}{c}{\textit{\textbf{mA AUC Pr-Re}}} &    & 0.991	& 0.994	& 0.930 & 	0.909       \\ \hline 
    
    \multicolumn{1}{c}{\textit{\textbf{mA AUC ROC}}}   &  WE & 0.993 &	0.985 &	0.977 &	0.959        \\ 
    \multicolumn{1}{c}{\textit{\textbf{mA AUC Pr-Re}}} &   & 0.979 &	0.943 &	0.902 &	0.854       \\ \hline
    
    \multicolumn{1}{c}{\textit{\textbf{mA AUC ROC}}}   &  WMV & 0.999 &	0.999 &	0.982 &	0.994  \\
    \multicolumn{1}{c}{\textit{\textbf{mA AUC Pr-Re}}} &  & 0.999 &	0.996 &	0.917 &	0.971    \\ \hline\hline
    
    \end{tabular}%
    }
    \caption{Sensor identification results.}
    \label{tbl:scores_of_texture_descriptor_methods}
    \end{center}
\end{table}%

For ROI data, results deteriorate slightly. In particular, AUC Pr-Re for spatial domain techniques is no longer acceptable. DFT and wavelet-based descriptors however still result in values well above 0.9, in most cases above 0.95, which is a very good result, that could not be expected given the high similarity of textures and the simplicity of our descriptors. For ROI data, there is no clear trend if enhancement as being applied is beneficial or not. In any case, similar to the original sample case, for well performing techniques the difference is negligible.




\section{\textbf{Conclusion}}
\label{sec:conclution}

We have identified simple texture descriptors as being well suited for finger vein sensor {\it model} identification, being applied to raw sample images as well as to the more challenging finger vein ROI data. Enhancement techniques turn out to be non-decisive for classification accuracy, at least when considering top performing techniques. Overall, but especially when considering results for ROI data, Fourier and wavelet-domain descriptors are found to perform superior to spatial domain techniques.

The excellent results suggest the proposed techniques to be better suited as compared to PRNU-based methods for the task investigated. Also, a fusion of both approaches seems promising, which will be subject to further investigations.




{\small
\bibliography{egbib}

\bibliographystyle{unsrt}
}

\end{document}